\documentclass[runningheads]{llncs}

\usepackage{float}
\usepackage{xcolor}
\usepackage{tcolorbox}
\usepackage{graphicx}
\usepackage{multirow}
\usepackage{booktabs}
\usepackage{bbding}
\usepackage{hyperref, color}

\usepackage[section]{placeins}

\usepackage{amsmath}
\usepackage{algorithm}
\usepackage{algorithmic}
\usepackage[switch]{lineno}
\usepackage{threeparttable}
\usepackage{rotating}

\definecolor{darkGreen}{RGB}{0,100,0}  
\definecolor{darkPurple}{RGB}{128,0,128}  

\begin{document}
\title{Can DeepSeek Reason Like a Surgeon? An Empirical Evaluation for Vision-Language Understanding in Robotic-Assisted Surgery}
\titlerunning{Can DeepSeek Reason Like a Surgeon?}
%
\author{Boyi Ma\inst{1} $^\star$
\and {Yanguang Zhao}\inst{1}
\thanks{Equal Contribution.}
\and Jie Wang\inst{1}
\and Guankun Wang\inst{1}
\and Kun Yuan\inst{2}
\and Tong Chen\inst{1}
\and Long Bai\inst{1}
\thanks{Project Lead.}
\and Hongliang Ren\inst{1} 
\thanks{Corresponding Author.}
}

\authorrunning{B. Ma et al.}
\institute{Department of Electronic Engineering, The Chinese University of Hong Kong
\and University of Strasbourg, CNRS, INSERM, ICube \& IHU Strasbourg \\
\email{b.long@link.cuhk.edu.hk, hlren@ee.cuhk.edu.hk}
}
\maketitle              
\begin{abstract}
The DeepSeek models have shown exceptional performance in general scene understanding, question-answering (QA), and text generation tasks, owing to their efficient training paradigm and strong reasoning capabilities. In this study, we investigate the dialogue capabilities of the DeepSeek model in robotic surgery scenarios, focusing on tasks such as Single Phrase QA, Visual QA, and Detailed Description. The Single Phrase QA tasks further include sub-tasks such as surgical instrument recognition, action understanding, and spatial position analysis. We conduct extensive evaluations using publicly available datasets, including EndoVis18 and CholecT50, along with their corresponding dialogue data. Our empirical study shows that, compared to existing general-purpose multimodal large language models, DeepSeek-VL2 performs better on complex understanding tasks in surgical scenes. Additionally, although DeepSeek-V3 is purely a language model, we find that when image tokens are directly inputted, the model demonstrates better performance on single-sentence QA tasks. However, overall, the DeepSeek models still fall short of meeting the clinical requirements for understanding surgical scenes. Under general prompts, DeepSeek models lack the ability to effectively analyze global surgical concepts and fail to provide detailed insights into surgical scenarios. Based on our observations, we argue that the DeepSeek models are not ready for vision-language tasks in surgical contexts without fine-tuning on surgery-specific datasets.
\end{abstract}
\section{Introduction}
\label{sec:introduction}

Robotic-assisted surgery (RAS) is a pivotal advancement in modern healthcare, aiming to assist surgery procedures and enable remote operation~\cite{fiorini2022concepts}. With the advancements in artificial intelligence and robotic control systems, contemporary RAS technology has demonstrated superior precision in control~\cite{huang2024endo,zhong2023integrated}. Though significantly in alleviating the workload of surgeons, current RAS heavily relies on expert surgeons to control the manipulator and issue precise commands~\cite{ding2025visual,marcus2024ideal}. 
Training a skilled surgeon takes significant time and effort, resulting in a shortage of highly qualified professionals~\cite{chen2024vs,seenivasan2023surgicalgpt}. Therefore, it is significant to develop a powerful AI model capable of comprehending and analyzing surgical scenarios for the potential substitution of expert surgeons.

A crucial aspect of intelligent RAS systems is their capability to understand complex surgical environments and provide context-aware assistance, where accurately analyzing visual and contextual information is essential~\cite{chen2024asi,psychogyios2023sar,wang2024video,zia2023surgical}. Building upon recent advancements in RAS, researchers have explored vision-language models to enhance the system’s ability to interpret surgical scenes and assist decision-making~\cite{chen2025text,shen2025medical,wang2023dynamic}.  
To this end, visual question answering (VQA) has emerged as a key technique, enabling AI models to extract task-relevant features by dynamically attending to critical regions within surgical images based on natural language queries~\cite{bai2025surgical,bai2023revisiting,wang2024surgical,yuan2024advancing}.  
The evolution of VQA methodologies in RAS has progressed from early long short-term memory (LSTM) networks~\cite{hochreiter1997long} to attention-based models~\cite{bai2023cat,vaswani2017attention,yuan2023learning}, and more recently, to transformer architectures~\cite{bai2023surgical,seenivasan2022surgical,zhou2023transformer}.  
Despite these advancements, existing VQA models still face challenges in seamlessly integrating multimodal perception, real-time comprehension, and efficient decision support, limiting their practical deployment.

Recent Multimodal Large Language Models (MLLMs)~\cite{jiang2023mistral7b,wang2024copesd,wang2024qwen2} have demonstrated significant advancements in their ability to interpret both visual and linguistic information. Leveraging vast amounts of training data, MLLMs surpass traditional VQA models by effectively addressing open-ended queries and providing context-aware responses. Given their remarkable success in general-purpose applications, MLLMs present new opportunities for enhancing contextual understanding and question-answering (QA) in the surgical domain~\cite{jiang2023mistral7b}. Recent studies have begun exploring the potential of MLLMs in surgical AI through supervised fine-tuning of open-source models~\cite{chen2024vs,hao2025enhancing,li2024llava,wang2024surgical,wang2025endochat}, aiming to adapt their capabilities to domain-specific challenges.  
The recently emerging open-source large language model, DeepSeek~\cite{wu2024deepseek,chen2025janus,liu2024deepseek} has gradually demonstrated superior performance compared to other open-source models, while utilizing fewer activated parameters.
Through comprehensive optimizations in the training framework, DeepSeek models achieve cost-efficient training and high inference efficiency. Moreover, leveraging reinforcement learning~\cite{guo2025deepseek}, it exhibits exceptional reasoning capabilities. Given its remarkable performance alongside cost-effective training and inference, DeepSeek models are poised to become a promising solution in surgical vision-language learning~\cite{lai2025med,pan2025medvlm}.  

Although DeepSeek models show exceptional performance in general tasks, their effectiveness in contextual understanding and reasoning within the RAS domain remains uncertain.
To bridge this gap, we systematically evaluate its potential using two publicly available robotic surgery datasets, EndoVis18~\cite{allan20202018} and CholecT50 datasets~\cite{nwoye2022data}. 
Specifically, we compare GPT-4o with open-source model DeepSeek-Janus-Pro-7b, DeepSeek-VL2, and DeepSeek-V3 in their ability to generate accurate and contextually relevant responses across three distinct dialogue paradigms: single-phrase QA, visual QA, and detailed descriptions. Through the above investigations, we aim to assess DeepSeek models' applicability to the RAS domain and provide actionable insights for optimizing its deployment in future surgical AI applications. 
Overall, the current DeepSeek models still struggle in comprehension of surgical targets and instruments, making them difficult for direct applications on surgical scene understanding tasks.
To develop surgical MLLMs capable of performing complex descriptions and reasoning, high-quality vision-language surgical datasets remain a critical necessity.

\section{Implementation}
We include GPT-4o~\cite{hurst2024gpt}, DeepSeek-Janus-Pro-7b~\cite{chen2025janus}, and DeepSeek-VL2~\cite{wu2024deepseek} as comparison models in our empirical study. Additionally, the language-based model DeepSeek-V3~\cite{liu2024deepseek} does not possess direct image reading capabilities. Therefore, we use the SEED tokenizer~\cite{ge2023making}, an image tokenization technique aligned with the pre-trained unCLIP Stable Diffusion~\cite{rombach2022high} to generate the image token. Given the image token, we specify an extra prompt to DeepSeek-V3 for its understanding: '\textit{You are given image tokens which you should interpret as an image}'.
Considering the varied dialogue demands of MLLM and the complexity of surgical procedures, the experiments are divided into three different paradigms: Single Phrase QA, Visual QA, and Detailed Description. These complementary paradigms enable the evaluation to provide more comprehensive and precise insights into the models' reactions toward surgical knowledge. Our evaluation utilizes three distinctive paradigms established by EndoChat dataset~\cite{wang2025endochat}, including surgical images from the MICCAI EndoVis18~\cite{allan20202018} and CholecT50 datasets~\cite{nwoye2022data}, while following the test set partitions specific to each dataset. 

\begin{figure*}[htbp]
\centering
\begin{minipage}{1.0\columnwidth}\vspace{0mm}    \centering
\begin{tcolorbox} 
    \centering
   
      \footnotesize
    \begin{tabular}{p{0.97\columnwidth} c}
 \textit{\textbf{Question [Single Phrase QA]:}} & \hspace{-4.0cm} \multirow{5}{*}{ \includegraphics[height=3.0cm]{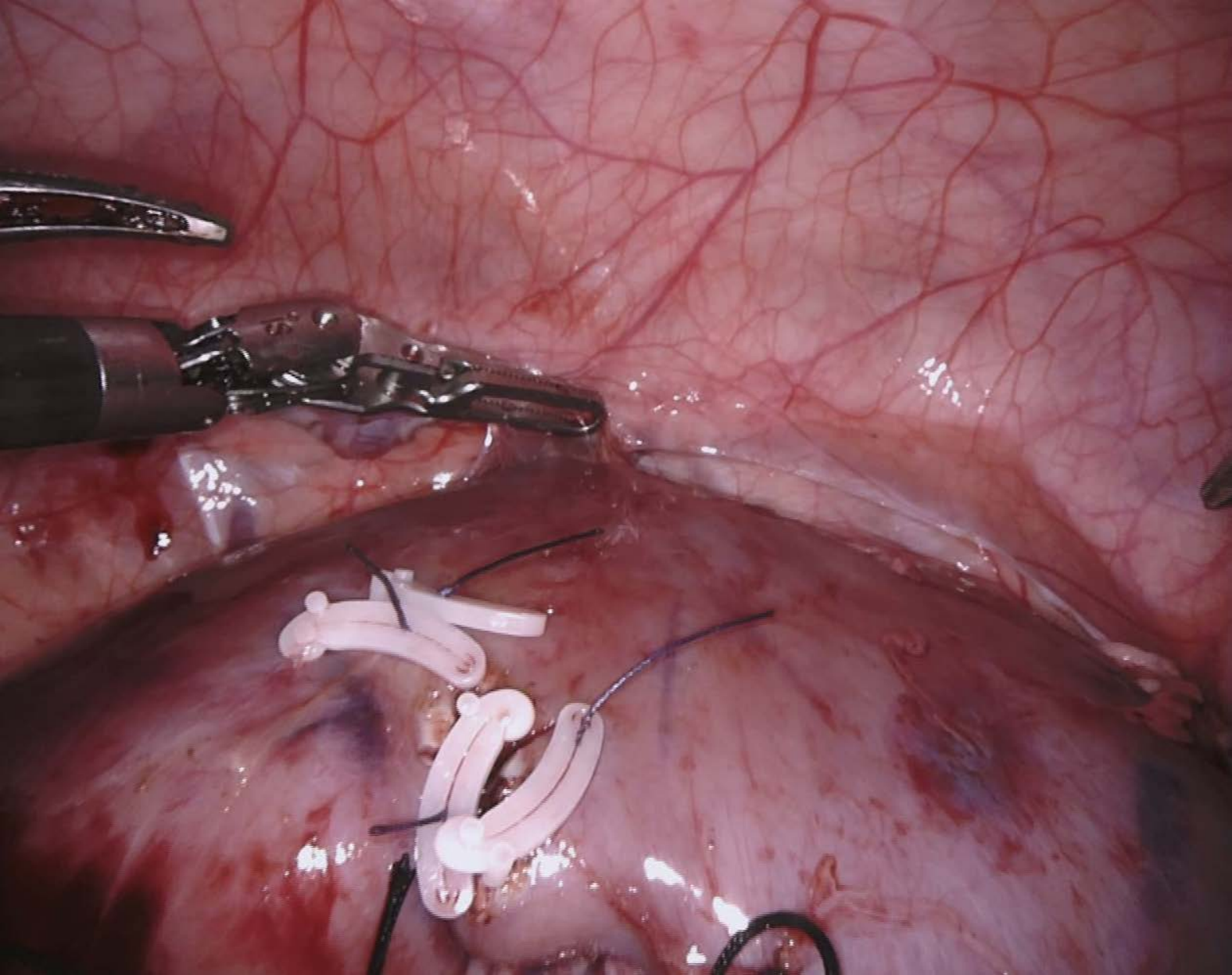} }
\\ 

What is the state of monopolar curved scissors? & \\ 
Choose one answer from ``Idle", ``Grasping", & \\ 
``Tool Manipulation", ``Cutting", ``Suction", & \\ 
``Retraction", ``Clipping", ``Cauterization", & \\
``Tissue Manipulation", ``Ultrasound Sensing",& \\ 
``kidney", ``left-top", ``left-bottom", ``right-top", & \\ 
``right-bottom". Only answer your choice.  & \\ 
    \textcolor{darkPurple}{\textbf{Ground Truth:}} Idle. & \\ 
    \textcolor{darkGreen}{\textbf{GPT-4o:}} \textcolor{red}{Cutting}. \hspace{6.6em}
   \textcolor{blue}{\textbf{DeepSeek-Janus-Pro-7b:}} \textcolor{red}{Cutting}. \\
   \textcolor{blue}{\textbf{DeepSeek-VL2:}} Idle. \hspace{5em}
   \textcolor{blue}{\textbf{DeepSeek-V3:}} Idle. & \\
    & \\

        \textit{\textcolor{black}{\textbf{Question [Single Phrase QA]:}}} What is the state of bipolar forceps? Choose one answer from ``Idle", ``Grasping", ``Tool Manipulation", ``Cutting", ``Suction", ``Retraction", ``Clipping", ``Cauterization", ``Tissue Manipulation", ``Ultrasound Sensing", ``kidney", ``left-top", ``left-bottom", ``right-top", ``right-bottom". Only answer your choice. & \\
            \textcolor{darkPurple}{\textbf{Ground Truth:}} Idle. & \\
    \textcolor{darkGreen}{\textbf{GPT-4o:}} \textcolor{red}{Cauterization}.  \hspace{4.0em}
   \textcolor{blue}{\textbf{DeepSeek-Janus-Pro-7b:}} Idle. & \\
   \textcolor{blue}{\textbf{DeepSeek-VL2:}} Idle.  \hspace{5em}
   \textcolor{blue}{\textbf{DeepSeek-V3:}} Idle. & \\
    & \\
   \textit{\textcolor{black}{\textbf{Question [Visual QA]:}}} What tools are operating the organ? & \\
\textcolor{darkPurple}{\textbf{Ground Truth:}} The tools operating are \underline{bipolar forceps}, \underline{prograsp forceps}, \underline{monopolar curved scissors}. & \\
  \textcolor{darkGreen}{\textbf{GPT-4o:}} Robotic \underline{surgical instruments} are operating the organ. & \\
   \textcolor{blue}{\textbf{DeepSeek-Janus-Pro-7b:}} \underline{Surgical instruments}. & \\
   \textcolor{blue}{\textbf{DeepSeek-VL2:}} The organ is operated by a combination of \textcolor{red}{mechanical and electronic components}.& \\
   \textcolor{blue}{\textbf{DeepSeek-V3:}} The tools operating the organ are \underline{surgical instruments} and robotic arms. & \\
& \\

   \textit{\textcolor{black}{\textbf{Question [Visual QA]:}}} What is the purpose of using prograsp forceps in this kind of surgery? & \\
\textcolor{darkPurple}{\textbf{Ground Truth:}} Prograsp forceps are used in minimally invasive robotic surgeries to \underline{grasp, manipulate, and retract tissues}. In this case, they appear to be \underline{retracting tissue} to provide better access and visibility to the kidney that is being operated on. & \\
  \textcolor{darkGreen}{\textbf{GPT-4o:}} Prograsp forceps are used for precise \underline{tissue manipulation} and \underline{traction} in minimally invasive surgery. & \\
   \textcolor{blue}{\textbf{DeepSeek-Janus-Pro-7b:}} To hold tissue securely during surgery. & \\
   \textcolor{blue}{\textbf{DeepSeek-VL2:}} Prograsp forceps are used to \underline{grasp and hold tissue} or organs during surgery, helping surgeons perform precise movements. & \\
   \textcolor{blue}{\textbf{DeepSeek-V3:}} The purpose of using prograsp forceps is to provide precise \underline{grasping and manipulation of tissues} during the surgery. & \\
    \end{tabular}
\end{tcolorbox}
\vspace{-2mm}
\caption{Qualitative examples of Single Phrase QA, Visual QA, and Detailed Description on the EndoVis18 dataset.}
    \label{fig:example_1}
\end{minipage}
\end{figure*}

Single Phrase QA focuses on providing concentrated and conclusive answers for specific details of surgical instruments, organs, actions, or relative positions. The model needs to react rapidly and accurately to specific visual content on surgical procedures without verbose elaboration. Through the utilization of the task-specific prompt: “\textit{Choose one answer from [list]}”, the task is streamlined to the process of identifying the correct answer through the understanding of visual content, as shown in Fig.~\ref{fig:example_1}. When concentrating on the Single Phrase QA task, the model adeptly links visual content with its corresponding textual annotations, which is critical for accomplishing subtasks.

Visual QA evaluates the capability of the model to mimic a surgeon expert to have a natural conversation. The MLLM is expected to provide concise and contextual clarity answers based on varying question formats, boosting its ability to act as a surgeon-like interaction model. General questions are provided as prompts, as shown in Fig.~\ref{fig:example_1}, such as “\textit{What tools are operating the organ?}” “\textit{Can you identify the instruments that are visible?}” “\textit{What type of procedure is depicted in the image?}”. Visual QA is crucial in addressing queries from surgeons and interns in diverse scenarios and enables the model to focus on delivering pertinent responses to different surgical scenes. 

\begin{figure*}[htbp]
\centering
\begin{minipage}{1.0\columnwidth}\vspace{0mm}    \centering
\begin{tcolorbox} 
    \centering
   
      \footnotesize
    \begin{tabular}{p{0.97\columnwidth} c}
  \textit{\textbf{Question [Detailed Description]:}} & \hspace{-4.0cm} \multirow{5}{*}{ \includegraphics[height=3.0cm]{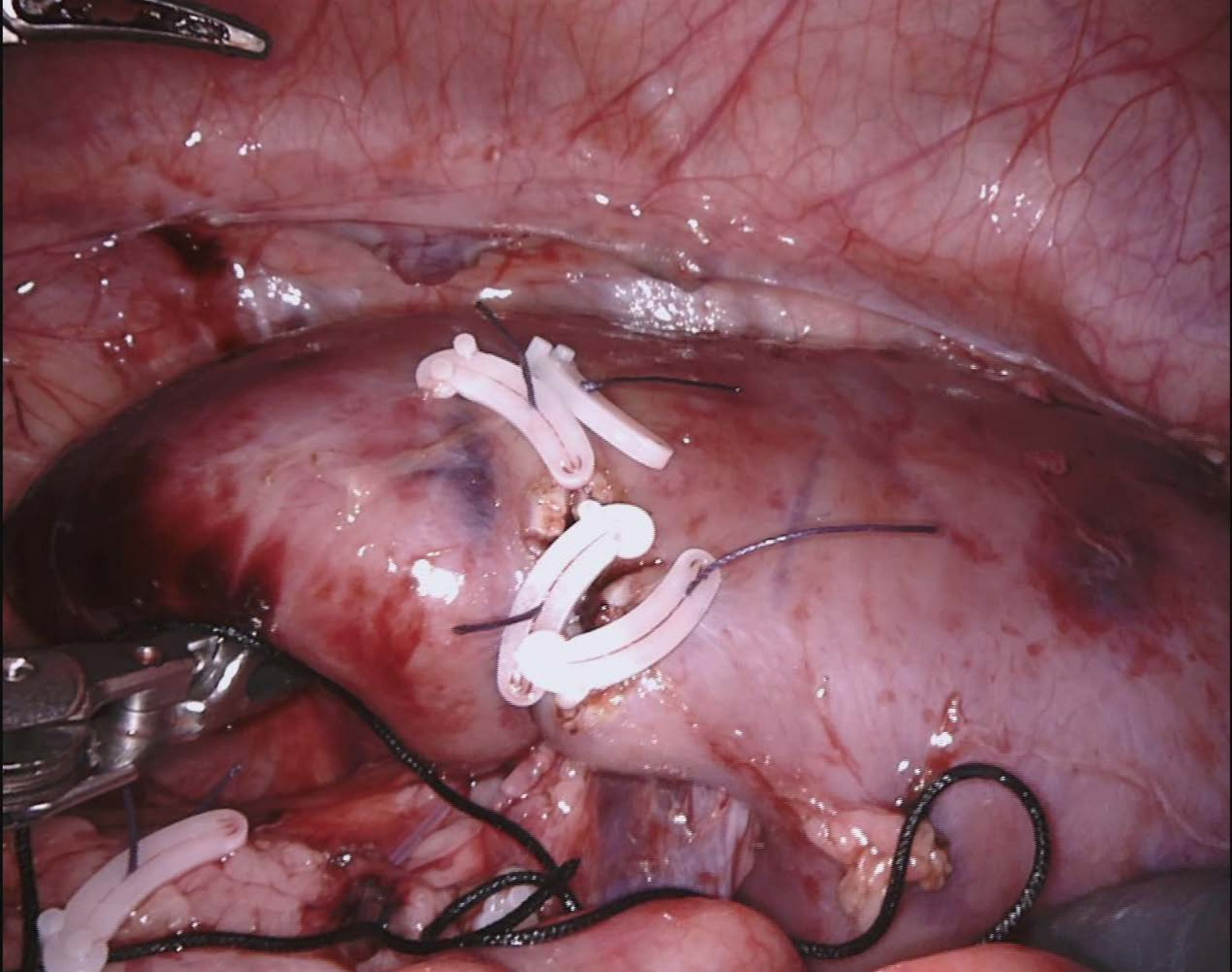} }
  \\
 Offer a thorough analysis of the image. & \\
    \textcolor{darkPurple}{\textbf{Ground Truth:}} The image depicts a surgical  & \\ 
    scene featuring a \underline{kidney}, with a \underline{reddish brown} & \\ 
    \underline{tone}  and smooth surface appearance. In relation & \\ 
    to the \underline{surgical instruments} present, there are  & \\ 
    \underline{bipolar forceps} located in the \underline{upper left} side of  & \\
    the image. They appear to be at rest and not & \\
    currently engaged in any activity. Towards the \underline{lower left} part of the image, the \underline{prograsp forceps} can be seen, which are being utilized for \underline{manipulating tissue}. The overall composition of the image with the kidney and surgical tools provides a representation of a \underline{complex medical procedure}. & \\
    \textcolor{darkGreen}{\textbf{GPT-4o:}} The image shows a \underline{medical procedure} involving an \underline{organ}, likely the liver, with visible \underline{surgical tools} and clips. The presence of sutures and clips suggest a repair or biopsy. The tissue appears inflamed, indicating recent \underline{surgical intervention}. The environment is consistent with an abdominal cavity, characterized by the glossy surface and vascular texture of the organ. The pale color of the clips and sutures contrasts with the \underline{darker, reddish tissues}, emphasizing their surgical purpose.. & \\
   \textcolor{blue}{\textbf{DeepSeek-Janus-Pro-7b:}} The image shows a \underline{surgical procedure} involving the insertion of medical devices into the body. The devices appear to be surgical clips or retractors, which are used to \underline{hold tissues} open during surgery. The \underline{surrounding tissue} is exposed, and the \underline{surgical instruments} are visible, indicating that the procedure is in progress. The patient's body is shown with a focus on the internal structures, likely for a specific medical or diagnostic purpose. The image highlights the complexity and precision required in surgical procedures to ensure patient safety and successful outcomes. & \\

  \textcolor{blue}{\textbf{DeepSeek-VL2:}} The image shows an intraoperative view of a \underline{surgical procedure} involving the heart. The presence of sutures and clips suggests that this is likely a cardiac surgery, possibly related to valve repair or replacement. The visible structures include the heart muscle and \underline{surrounding tissues}, with \underline{instruments} in use for the procedure. The coloration indicates the presence of blood, which is typical during such surgeries. The overall setting implies a controlled environment, likely within an operating room. & \\
   \textcolor{blue}{\textbf{DeepSeek-V3:}} The image appears to be a complex arrangement of recurring elements, suggesting a structured yet \textcolor{red}{dynamic composition}. The composition seems balanced, with \textcolor{red}{a mix of larger and smaller components} distributed evenly across the frame. The overall tone suggests a blend of order and creativity, possibly \textcolor{red}{representing a harmonious design or an abstract representation of interconnected ideas}. The interplay of recurring motifs and subtle variations could evoke \textcolor{red}{themes of unity, repetition, or evolution}, depending on the viewer's interpretation. & \\

    \end{tabular}
\end{tcolorbox}
\vspace{-2mm}
\caption{Qualitative examples of Single Phrase QA, Visual QA, and Detailed Description on the EndoVis18 dataset.}
    \label{fig:example_2}
\end{minipage}
\end{figure*}

Detailed Description requires the model to gain a comprehensive understanding of organs, instruments, actions, relative positions, and contextual information at the same time. It’s necessary to have a detailed grasp and overview of the entire scope to establish the foundation for more complex surgical procedures. By giving prompts such as “\textit{Describe the following image in detail}” and “\textit{Clarify the contents of the displayed image with great detail}”, the model’s ability to provide local and global information about the surgical scenario is evaluated.

\section{Results and Analysis}
\subsection{Performance on Single Phrase QA}

\begin{table}[t]
    \centering
    \caption{Quantitative comparison of the Single Phrase QA on the EndoVis18 and CholecT50 datasets.}
    \resizebox{\textwidth}{!}{
    \begin{tabular}{cccccccccccc}
    \toprule
    \multirow{2}[2]{*}{Model} & \multirow{2}[2]{*}{Dataset} &\multicolumn{2}{c}{Average}&\multicolumn{2}{c}{Instrument Motion}&\multicolumn{2}{c}{Target Issue}&\multicolumn{2}{c}{Object Position}\\
    \cmidrule{3-10}&  &  Accuracy &  F1 Score&  Accuracy &  F1 Score&  Accuracy &  F1 Score&  Accuracy &  F1 Score\\
    \midrule
    {GPT-4o}& \multirow{4}{*}{EndoVis18}& 0.105 & 0.032& 0.138 & 0.109& 0.251 & 0.013& 0.016 & 0.003\\
    {DeepSeek-Janus-Pro-7B} & & 0.187 & 0.068 & 0.447 & 0.128& 0.000 & 0.000 & 0.000 & 0.000 \\
    {DeepSeek-VL2} & & 0.137 & 0.115 & 0.223 & 0.205 & 0.000 & 0.000& 0.104 & 0.035\\
    {DeepSeek-V3} & & 0.276 & 0.109 & 0.276 & 0.166 & 0.996 & 0.333 & 0.000 & 0.000 \\
    \midrule
    \multirow{2}[2]{*}{Model}& \multirow{2}[2]{*}{Dataset} &\multicolumn{2}{c}{Average}&\multicolumn{2}{c}{Instrument}&\multicolumn{2}{c}{Target}&\multicolumn{2}{c}{Motion}\\  
    \cmidrule{3-10}&  &  Accuracy &  F1 Score&  Accuracy &  F1 Score&  Accuracy &  F1 Score&  Accuracy &  F1 Score\\
    \midrule
    {GPT-4o}& \multirow{4}{*}{CholecT50}& 0.592 &0.376& 0.873 & 0.846& 0.336 & 0.167& 0.566 & 0.434\\
    {DeepSeek-Janus-Pro-7B} &  & 0.250 & 0.148 & 0.302 & 0.324 & 0.115 & 0.042 & 0.333 & 0.195\\
    {DeepSeek-VL2} & & 0.349& 0.216 & 0.543 & 0.446 & 0.111 & 0.019 & 0.394 & 0.324\\
    {DeepSeek-V3}& &0.573 & 0.404& 0.860 & 0.817& 0.395 & 0.210& 0.464 & 0.386\\
    \bottomrule
    \end{tabular}}
    \label{tab:singleQA}
\end{table}
As shown in Table~\ref{tab:singleQA}, accuracy and F1 score are chosen to evaluate the performance on the Single Phrase QA task. The variable choices from the EndoVis18 dataset contain 13 different phrases covering surgical instrument motion, target issue, and object position, encompassing 8, 1, and 4 phrases, respectively. The phrases from the CholecT50 dataset are separated into three categories: surgical instrument, targets, and motion, which contain 5, 8, and 13 phrases, respectively. 

Table~\ref{tab:singleQA} shows that the DeepSeek models have overall better results on the EndoVis18 dataset. The performance in the instrument motion category is substandard, which shows that both DeepSeek and GPT4-o face challenges in deducing the instrument motion from a static image. Additionally, we observed that DeepSeek-Janus-Pro-7b predominantly selects the `Idle' category across all motion classifications. The `Idle' label happens to be the most frequently reported outcome, thereby the overall results appear higher than other models. However, this does not mean that DeepSeek-Janus-Pro-7b possesses a superior capability in classifying instrument motion. Regarding the target issue category, the options provided in the prompt include just a single choice associated with the target issue, namely, kidney. Therefore, within this category, selecting the correct option is straightforward once the model comprehends both the question and the choices presented in the prompt. Based on the comparative outcomes, it is evident that DeepSeek-V3 demonstrates a superior grasp of the question and the provided information within the prompt. This superior performance can be attributed to DeepSeek-V3's capacity to generate outputs through reasoned responses to the questions rather than relying on random guessing for answers. However, both DeepSeek-Janus-Pro-7b and DeepSeek-VL2 exhibit poor performance, indicating that these models not only lack sufficient knowledge of surgical scenes but also struggle to comprehend the provided prompts effectively. 
Regarding the object position task, the model is expected to initially analyze and grasp the mentioned object and subsequently determine the object's position. The reasoning process of object position should begin with a general understanding of surgical phases, followed by the recognition of surgical instruments and tissues. Combining this knowledge with image information, the model could accurately locate the positions of relevant objects. While this task is straightforward for an expert surgeon, it poses significant challenges for GPT-4o and the DeepSeek series, primarily due to its limited domain-specific knowledge in robotic surgery.

Moreover, on the CholecT50 dataset, DeepSeek-V3 also demonstrates comparable performance with GPT-4o. The evaluation of various phrases reveals that analyzing surgical tissue poses the greatest challenge for GPT4-o and the DeepSeek series. In comparison, surgical instruments, such as scissors, clippers, and hooks, are more readily identifiable. This is likely due to the greater frequency of daily used tools in the MLLM training data compared to surgical tissues. While DeepSeek-V3 exhibits performance comparable to GPT-4o, both DeepSeek-Janus-Pro-7b and DeepSeek-VL2 yield poor results. We also observe that DeepSeek-VL2 tends not to follow the instruction '\textit{Only answer your choice},' sometimes providing lengthy explanations for its selection. These shortcomings indicate that DeepSeek-Janus-Pro-7b and DeepSeek-VL2 not only struggle to understand surgical scenes but may also face issues with following instructions. In conclusion, despite the existing limitations of MLLMs in achieving a comprehensive understanding of surgical scenarios involving instruments, tissues, and motions, DeepSeek-V3 demonstrates performance that is comparable to or even superior in simple QA tasks within surgical contexts. Although DeepSeek-V3 is inherently a language model without built-in visual processing capabilities, we are surprised to observe that, through our approach of tokenizing images, DeepSeek-V3 achieves favorable results in simple QA tasks related to surgical scenarios. This performance may be attributed to the internal data used during the training of DeepSeek-V3, although the specific reasons remain unclear.

\subsection{Performance on Visual QA}

\begin{table}[t]
    \centering
    \caption{Quantitative comparison of Visual QA and Detailed Descriptions on EndoVis18 datasets.}
    \resizebox{\textwidth}{!}{
    \begin{tabular}{cccccccc}
    \toprule
         \multirow{2}[2]{*}{Model} & \multicolumn{7}{c}{Visual QA}\\
         \cmidrule{2-8} & BLEU-3  &  BLEU-4  & CIDEr  &  METEOR  &  ROUGE-L  & ROUGE-1 & GPT Score\\
         \midrule
         GPT-4o & 0.127 & 0.099 & 0.627 & 0.157 & 0.292 & 0.369& 2.928\\
         DeepSeek-Janus-Pro-7B & 0.003 & 0.003 & 0.069 & 0.048 & 0.042 & 0.045 & 2.616 \\
         DeepSeek-VL2 & 0.134 & 0.108 & 0.514 & 0.142 & 0.231 & 0.296 & 2.006\\
         DeepSeek-V3 & 0.209& 0.158 & 0.892 & 0.217 & 0.461 &0.545  & 2.550\\
         \midrule
         \multirow{2}[2]{*}{Model}& \multicolumn{7}{c}{Detailed Description}\\
         \cmidrule{2-8} & BLEU-3&  BLEU-4& CIDEr& METEOR&  ROUGE-L& ROUGE-1 & GPT Score\\
         \midrule
         GPT-4o & 0.042 & 0.022 & 0.007 & 0.135 & 0.161 &  0.231& 1.154\\
         DeepSeek-Janus-Pro-7B & 0.057 & 0.032 & 0.007 & 0.141 & 0.181 & 0.235 & 0.891\\
         DeepSeek-VL2 & 0.047& 0.025 & 0.010 & 0.134 & 0.162 & 0.259 & 1.156\\
         DeepSeek-V3 & 0.023 & 0.010 & 0.003 & 0.118 & 0.153 & 0.169 & 0.213\\
    \bottomrule
    \end{tabular}}
    \label{tab:visualQA}
\end{table}
The evaluation results of Visual QA on the EndoVis18 dataset are demonstrated in Table~\ref{tab:visualQA}. The performance of generated responses is evaluated across sequences, quality, and consensus utilizing six different metrics. Additionally, the GPT-3.5 score is employed to determine whether the generated information is accurately mentioned or partially matched. In Visual QA tasks, the model is expected to provide a generally accurate understanding of surgical contexts. Therefore, the focus should be on semantic relevance and basic lexical coverage, particularly as measured by CIDEr and ROUGE-1 metrics. As shown in Table~\ref{tab:visualQA},  despite DeepSeek-V3 being a language-based model, it demonstrates superior performance in both CIDEr and ROUGE-1 by feeding image tokens into the LLM directly. With appropriately designed guiding prompts, DeepSeek-V3 shows the ability to understand the posed questions and generate surgical-related lexical content based on its image perception capabilities. Furthermore, the consistent performance across six different metrics of DeepSeek-VL2 and DeepSeek-V3 highlights their efficiency in accurately capturing key concepts and focusing on the relevant visual content. Through our evaluation, we also observe that DeepSeek-Janus-Pro-7b tends to provide short answers, even when we include `\textit{Answer with short sentences}' in the prompt. This indicates its limited instruction following ability and insufficient understanding of surgical knowledge.

\subsection{Performance on Detailed Description}

When evaluating the Detailed Description task, we provide general prompts such as “\textit{Describe the following image in detail}” and “\textit{Clarify the contents of the displayed image with great detail}.” As shown in Fig.~\ref{fig:example_1} and Fig.~\ref{fig:example_2}, the ground truth demonstrates that the model is expected to produce a comprehensive description of the surgical scene. This involves identifying surgical instruments, describing operational motions, specifying the target tissue, and providing locational context. From Table~\ref{tab:visualQA}, it can be observed that all the models perform poorly on BLEU and CIDEr metrics. The low performance can be attributed to the excessive length of the comparison sentences, which makes it challenging for the models to maintain n-gram precision and achieve reasonable similarity in this task. Since the Detailed Description task typically generates longer text, the order of the output content and the textual expression style can significantly impact its evaluation score. However, its generated text consistently includes details about instruments, motions, and tissues within the surgical context. 

As shown in Fig.~\ref{fig:example_2}, DeepSeek-V3 has limited ability to provide a general description, as DeepSeek-V3 is only a language model. In our previous tests, we achieved good performance in simple QA tasks by directly inputting image tokens. However, such a language model struggles to provide meaningful explanations when performing detailed description tasks. On the other hand, DeepSeek-VL2 and DeepSeek-Janus-Pro-7b demonstrate superior performance across multiple indicators compared to GPT-4o, showcasing their ability to capture the overall context of the image and provide detailed surgical descriptions, even when given a simple prompt. From Fig.~\ref{fig:example_2} it is evident that the text generated by DeepSeek-VL2 and DeepSeek-Janus-Pro-7b is not verbose; instead, it tends to provide concise summaries while offering more accurate content. This indicates that both models possess great capabilities in understanding and summarizing images. However, their lack of surgical scene comprehension limits their applicability in this domain. In conclusion, DeepSeek-VL2 can achieve reasonable results in detailed description tasks; however, due to its lack of surgical knowledge, the model cannot provide accurate answers and fails to include all the necessary details required in surgical scenarios.

\section{Conclusion}\label{sec:conclusion}
In this paper, we conduct empirical studies on the DeepSeek series to evaluate its contextual comprehension capabilities in robotic surgery. Our experiments focus on three distinct tasks: Single Phrase QA, Visual QA, and Detailed Description. The performance of DeepSeek-Janus-Pro-7b on three tasks shows its limitation in surgical scene comprehension and instruction understanding. DeepSeek-VL2, on the other hand, demonstrates a certain capacity for object position detection in Single Phrase QA tasks and can yield better results by providing concise summaries in detailed description tasks. However, due to its limited knowledge of surgical-related information and lack of comprehension of surgical scenes, DeepSeek-VL2 is not suitable for real-world surgical applications. Although DeepSeek-V3 is purely a language model, we test its performance by directly inputting image tokens and find that it performs well in single-sentence QA tasks. However, it fails to achieve a comprehensive understanding of surgical scenarios. From our experiments, we observe that DeepSeek-V3 can simulate basic surgical knowledge similar to that of a novice surgeon when given specific prompts. However, it struggles to interpret surgical actions and cannot accurately localize objects, which limits its practical clinical applications.

Based on the current results, there are still areas that require further improvement in future work: (i) Developing effective prompts is critical, as the clarity and specificity of instructions directly influence the model's performance. Crafting versatile and adaptable prompts can also enhance its ability to handle real-world interactive tasks. (ii) Integrating large-scale surgical datasets with reasoning-based fine-tuning can equip MLLMs with stronger inferential capabilities in surgical scenarios, enabling more effective collaboration between surgeons and the model. (iii) By incorporating strategies such as Mixture of Experts (MoE) and model distillation, computational and deployment costs for surgical MLLMs can be significantly reduced. Lower costs would support the development of embodied surgical robots and facilitate broader clinical adoption.

\subsubsection{Acknowledgements.}
This document is the results of the research projects by Hong Kong Research Grants Council (RGC) Research Impact Fund (RIF) R4020-22, Collaborative Research Fund (CRF) C4026-21GF, General Research Fund (GRF) 14203323, 14216022, and 14211420.

\bibliographystyle{splncs04}
\bibliography{references}

\end{document}